\documentclass[11pt]{article}

\usepackage[final]{acl}

\usepackage{times}
\usepackage{latexsym}
\usepackage[T1]{fontenc}
\usepackage[utf8]{inputenc}

\usepackage{microtype}

\usepackage{inconsolata}

\usepackage{graphicx}

\usepackage{booktabs}
\usepackage{algorithm}
\usepackage{algpseudocode}
\usepackage{multirow}
\usepackage{amsmath,amssymb,amsfonts}
\usepackage{xspace}
\newcommand{\ourmethod}{\textsc{ClusterFusion}\xspace}

\makeatletter
\newif\ifsubmission
\ifacl@finalcopy\submissionfalse\else\submissiontrue\fi
\makeatother

\definecolor{aclblue}{HTML}{000099}
\hypersetup{colorlinks=true, citecolor=aclblue, linkcolor=aclblue, urlcolor=aclblue}

\title{\ourmethod: Hybrid Clustering with Embedding Guidance and LLM Adaptation}

\author{
  Yiming Xu$^{1}$  \quad
  Yuan Yuan$^{1}$ \quad
  Vijay Viswanathan$^{2}$ \quad
  Graham Neubig$^{2}$ \\
  $^{1}$Adobe \quad $^{2}$Carnegie Mellon University \\
  \texttt{\{yimxu, yuyuan\}@adobe.com} \\
  \texttt{\{vijayv, gneubig\}@andrew.cmu.edu}
}

\begin{document}
\maketitle

\begin{abstract}
The goal of text clustering is to organize and understand a corpus, where the meaning of each cluster matters as much as its membership. Conventional methods focus on partitioning a pre-trained embedding space geometrically, and leave the interpretation of each cluster as a post-hoc consideration. We invert this design and make a large language model the core of clustering: \ourmethod extracts explicit topic definitions and assigns each record by conditioning on them, so its clusters are interpretable by construction. It also adapts to new domains by supplying domain knowledge and user preferences through prompting, without the costly fine-tuning that conventional methods require. We further show that the quality of the extracted topics is governed by \emph{context organization}: the exemplars shown to the model and the order in which they appear, where reordering the same inputs alone can shift accuracy by up to 22\%, an effect we attribute to attention degradation over long contexts. Across three public benchmarks and two new domain-specific datasets, \ourmethod achieves state-of-the-art accuracy, with pronounced gains in specialized domains and substantially more accurate cluster interpretation than post-hoc labeling, without any training. To support future work, we release our newly constructed dataset and results on all benchmarks.%
\ifsubmission
\footnote{[Dataset and code link hidden for anonymous review.]}
\else
\footnote{\url{https://github.com/YimingXu1213/clusterFusion/}}
\fi
\end{abstract}

\section{Introduction}
Text clustering has long been framed as partitioning a vector space into compact groups. Elegant as this geometric view is, it overlooks why we need clustering: to organize and understand a corpus in service of a downstream goal, whether exploring patterns in data \citep{Johnson2023WhereDM} or pseudo-labeling a corpus for a downstream task such as retrieval \citep{Charikar1997IncrementalCA}. This goal-oriented nature imposes two requirements. First, the method must be conditioned on the goal, since the task is underspecified without knowing what the user seeks and which aspects of the data matter to them \citep{Caruana-2013}. Second, its output must be interpretable: what each cluster means should be as clear as which records it contains. \textbf{In this paper, we describe an algorithm that (1) flexibly adapts to a domain by conditioning on domain knowledge and user preferences, and (2) produces clusters that are inherently interpretable and causally related to their associated topics.}

Text clustering today is typically done via a two-step process of extracting embeddings from a dataset \citep{muennighoff2022mteb, wang2022text} and applying traditional clustering algorithms \citep{MacQueen1967, day1984efficient}, yet this process satisfies neither requirement. The resulting clusters are geometric regions whose meaning must be recovered post hoc, and conditioning them on a new goal or domain demands costly fine-tuning of the embedder, or semi-supervised feedback whose annotation cost is impractical at scale \citep{basu2002semi, basu2004active, bae2020interactive}.

Large language models (LLMs) offer a natural means to condition clustering on a goal stated in language, and a growing body of work applies them to clustering \citep{de2023idas, zhang2023clusterllm, wang2023goal, viswanathan2024large, feng2024llmedgerefine, jo-etal-2025-zerodl}. In most of these methods, however, embedding-space clustering remains the primary driver of the output and the LLM is confined to an auxiliary role, enriching representations or refining boundaries post hoc. The cluster's interpretation thus remains opaque.

A nascent line of work instead generates clusters end-to-end with an LLM \citep{huang2025text}, decomposing clustering into two stages: (1) \emph{topic extraction}, where the LLM proposes candidate topics from the corpus, and (2) \emph{topic assignment}, where it assigns each record to one of them. As currently realized, this paradigm faces three limitations that bear directly on the accuracy and interpretability of the resulting clusters. First, despite the LLM's central role, the organization of its input---which records it is shown, and in what order---has not been studied systematically, even though it strongly affects whether the extracted topics are stable and coherent. Second, records are assigned to bare topic names without explicit definitions, leaving the clusters neither reliably interpretable nor stably assigned, an ambiguity this line of work itself notes. Third, there is no systematic mechanism for incorporating domain knowledge or user preferences, leaving the LLM's central advantage for goal conditioning untapped.

We realize this design in \ourmethod, which places the LLM at the core of clustering: it extracts explicit topic definitions from the corpus and assigns each record by conditioning on them. Because a cluster's members then share an explicit, human-readable description, and it can further incorporate user preferences and domain knowledge through prompting, \ourmethod can explicitly target a downstream goal or application. We make the following contributions:

\begin{itemize}
  \item \textbf{A natively interpretable formulation of clustering.} Every cluster assignment is conditioned exclusively on explicit topic definitions mined from the dataset; the definition that describes a cluster is therefore the sole basis of its membership, a property termed \emph{faithfulness} by \citet{jacovi-goldberg-2020-towards}. Unlike post-hoc interpretation of embedding-based clustering, our proposed algorithm is interpretable by construction.
  \item \textbf{A localization of the task's difficulty, and the role of context organization.} Decomposing clustering into topic extraction and assignment, we find that assignment is near-saturated, whereas topic extraction is the binding constraint on cluster quality. We trace this bottleneck to \emph{context organization}, the selection and ordering of the exemplars shown to the model: holding model, prompt, and sample fixed, ordering alone accounts for up to a 22\% accuracy difference, which we attribute to attention degradation over long contexts and upper-bound with an oracle ordering.
  \item \textbf{Controllable domain adaptation without training.} We show that domain knowledge and user preferences can be supplied through prompting to steer both topic extraction and assignment, yielding large gains in specialized domains where fine-tuning is impractical (e.g., accuracy on OpenAI Codex rises from 53.8 to 66.0). This is a form of control that embedding-based methods do not offer.
  \item \textbf{State-of-the-art results and new resources.} We instantiate these findings in \ourmethod, a training-free method that attains the best accuracy across five datasets and yields substantially more accurate cluster descriptions than embedding-based labeling. We release a new human-annotated domain-specific benchmark (OpenAI Codex), our prompt templates, and results on all benchmarks.
\end{itemize}

\section{\ourmethod framework}
\ourmethod treats the LLM as the clustering core by decomposing clustering into two stages: (1) topic extraction, where the LLM extracts candidate topics with their definitions from the corpus, and (2) topic assignment, where each record is assigned to one of the proposed topics. A fundamental bottleneck lies in the limited context window, which is particularly restrictive for topic extraction on large datasets. To mitigate this, we introduce an embedding-guided subset partition stage that selects and organizes the input via \textit{sampling} and \textit{sorting}, described below.

\begin{figure*}[t]
  \centering
  \includegraphics[width=1.0\textwidth]{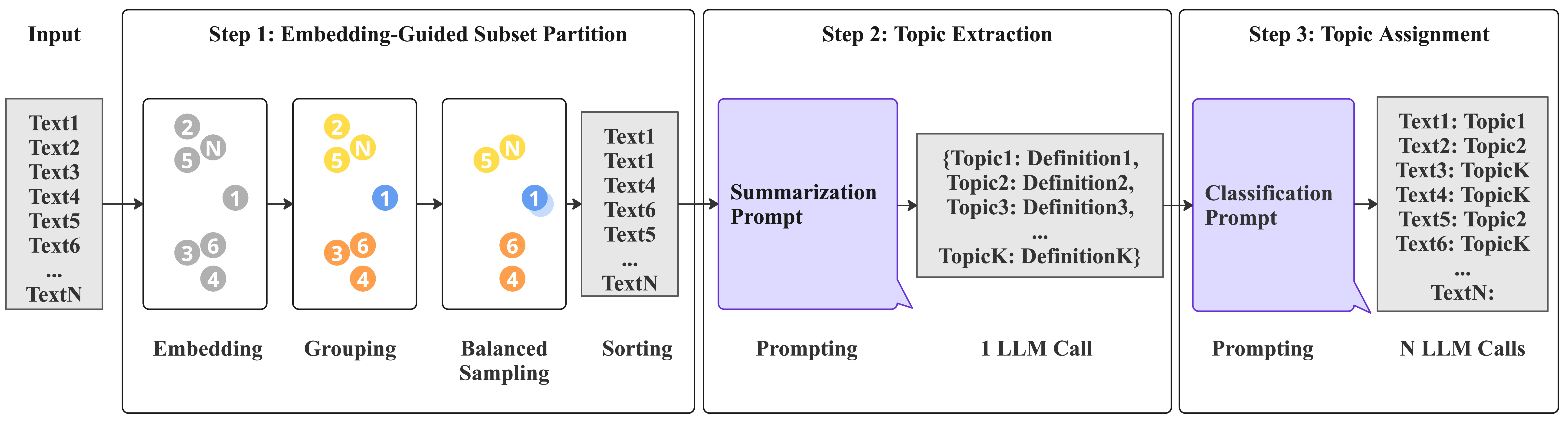}
  \caption{Overview of the \ourmethod framework}
  \label{fig:framework}
\end{figure*}

An overview of the framework is shown in Figure~\ref{fig:framework}, and the pseudocode is provided in Algorithm~\ref{alg:clusterfusion}.

\subsection{Embedding-guided subset partition}
We leverage embedding-based methods to transform the input dataset into a manageable, representative, and well-organized sample. This stage consists of two steps:

\paragraph{Grouping and balanced sampling.}
Given a dataset $D_N = \{x_1, x_2, \ldots, x_N\}$ with $N$ records, we first compute embedding representations:
\begin{equation}
Z = \{z_1, z_2, \ldots, z_N\} = \text{Embedder}(D_N)
\end{equation}
where $z_i \in \mathbb{R}^d$ is the embedding vector for record $x_i$.

We then apply a traditional clustering algorithm such as KMeans to partition the embeddings into $M$ groups:
\begin{equation}
G = \{G_1, G_2, \ldots, G_M\} = \text{Group}(Z, M)
\end{equation}
where each group $G_m = \{(x_j, z_j) : j \in \mathcal{I}_m\}$ contains records and their embeddings assigned to group $m$.

To construct a representative sample $D_S$ of size $S$, we perform balanced sampling by drawing $\lfloor S/M \rfloor$ samples from each group. For groups with sufficient samples ($|G_m| \geq \lfloor S/M \rfloor$), we sample without replacement to ensure diversity. For smaller groups ($|G_m| < \lfloor S/M \rfloor$), we sample with replacement to meet the target sample size:
\begin{equation}
D_S = \bigcup_{m=1}^{M} \text{Sample}(G_m, \lfloor S/M \rfloor)
\end{equation}
Sampling is necessary to respect the LLM's context window limit, while balanced sampling amplifies the signal from low-frequency clusters and ensures that minority groups remain represented.

\paragraph{Sorting.}
To improve coherence in the LLM input, we arrange the sampled records $D_S = \{x_{s_1}, x_{s_2}, \ldots, x_{s_S}\}$ before concatenation. We evaluate two sorting strategies:

\textit{(i) Cluster-based ordering:} Records are sorted by their predicted group index:
\begin{equation}
\pi_{\text{cluster}} = \arg\text{sort}(\{\text{group}(x_{s_i})\}_{i=1}^{S})
\end{equation}

\textit{(ii) Similarity-based ordering:} Records are sorted by their embedding similarity (cosine or Euclidean distance) to the first sampled record:
\begin{equation}
\pi_{\text{sim}} = \arg\text{sort}(\{\text{sim}(z_{s_1}, z_{s_i})\}_{i=1}^{S})
\end{equation}
where $\text{sim}(\cdot, \cdot)$ denotes a similarity measure (e.g., cosine similarity or Euclidean distance).

The final sorted sample is then $D_S' = \{x_{s_{\pi(1)}}, x_{s_{\pi(2)}}, \ldots, x_{s_{\pi(S)}}\}$, where $\pi \in \{\pi_{\text{cluster}}, \pi_{\text{sim}}\}$. Our experiments demonstrate that this pre-organization improves readability and reduces fragmentation, enabling the LLM to generate more consistent and higher-quality topics.

\subsection{Summarization: topic extraction}

The sampled and sorted records are concatenated into a prompt for the LLM, which is instructed to extract a compact set of candidate topics along with concise definitions. These topics serve as the proposed cluster labels. This step leverages the LLM's strong contextual reasoning abilities, and also enables the flexible incorporation of domain knowledge or user preferences directly into the prompt (e.g., specifying whether ambiguous utterances should be merged into one category or separated into distinct clusters). The complete prompt templates and domain-specific examples are provided in Appendix~\ref{appendix:prompts}.

\subsection{Classification: topic assignment}
Each record in the original dataset is then independently assigned to one of the candidate topics through a separate LLM call. The model is constrained to choose from the topic list produced in the \textit{Topic Extraction} step; if it outputs an invalid label, the call will be rerun. The full prompt template for this classification step is also detailed in Appendix~\ref{appendix:prompts}.

\begin{algorithm}[t]
\small
\caption{\ourmethod framework}
\label{alg:clusterfusion}
\begin{algorithmic}[1]
\item[] \textbf{Input:} Dataset $D_N$; number of clusters $K$
\item[] \textbf{Hyperparameters:} Number of groups $M$; total sample size $S$
\item[] \textbf{Output:} Cluster assignments for all $x \in D_N$
\vspace{0.5em}

\State \textbf{Step 1: Embedding-Guided Subset Partition}
\State $Z \gets \text{Embedder}(D_N)$
\State $G \gets \text{Group}(Z, M)$
\State $D_S \gets \emptyset$
\For{$m = 1 \to M$}
  \State $\text{replace\_flag} \gets \text{True if } |G_m| < S/M \text{ else False}$
  \State $D_S \gets D_S \cup \text{Sample}(G_m, S/M, \text{replace\_flag})$
\EndFor
\State $D_S \gets \text{Sort}(D_S)$

\State \textbf{Step 2: Topic Extraction}
\State $T \gets \text{LLM.extract}(D_S, K)$

\State \textbf{Step 3: Topic Assignment}
\ForAll{$x \in D_N$}
  \State $t \gets \text{LLM.classify}(x, T)$
  \State assign $x \mapsto t$
\EndFor

\State \Return clustering $\{x \mapsto t\}_{x \in D_N}$
\end{algorithmic}
\end{algorithm}

\section{Experimental setting}
\subsection{Datasets}
We evaluate \ourmethod on three widely used clustering benchmarks and two newly constructed domain-specific datasets. The generic benchmarks allow comparison with prior work, while the domain-specific corpora test the model's ability to adapt to specialized contexts. The existing generic benchmarks are:
\begin{itemize}
    \item \textbf{Bank77} \citep{casanueva2020efficient}: The Bank77 dataset contains 3,080 user queries directed to an online banking assistant, annotated into 77 intent categories.
    \item \textbf{CLINC} \citep{larson2019evaluation}: The CLINC dataset contains 4,500 user queries from a task-oriented dialog with 150 intent classes, after removing ``out-of-scope'' queries \citep{zhang2023clusterllm}. In addition, we found that a small amount of labels were ambiguous or inconsistent, and we performed further label cleaning.%
\ifsubmission
\footnote{Correction notes are available at: [link hidden for anonymous review].}
\else
\footnote{Correction notes are available at: \url{https://github.com/YimingXu1213/clusterFusion/blob/main/datasets/clinicSmall_correction_notes.csv}.}
\fi

    \item \textbf{Tweet} \citep{yin2016model}: The Tweet dataset contains 2,472 social media posts annotated into 89 intent categories.
\end{itemize}
New domain-specific benchmarks:\footnote{During data cleaning, we removed all personally identifiable information (PII), including usernames, profile links, and user IDs, as well as hateful or offensive content for both datasets. These steps ensure compliance with ethical standards.}
\begin{itemize}
    \item \textbf{OpenAI Codex}: We constructed a new dataset of 406 YouTube comments from the OpenAI Codex announcement video\footnote{\url{https://www.youtube.com/watch?v=FUq9qRwrDrI}} before June, 2025. The dataset reflects a highly technical domain and contains emerging terminology. Comments were manually annotated into 26 categories capturing user reactions, technical questions, and broader discussions.
    \item \textbf{Adobe Lightroom}: We collected 6,410 customer reviews for Adobe Lightroom iOS, annotated into 40 categories. This dataset captures domain-specific terminology and nuanced user feedback.
\ifsubmission
    The reviews were collected from publicly available App Store listings. Due to the platform's terms of service, we are unable to redistribute the raw data. However, we include our evaluation results on this dataset.
\else
    Due to Adobe company policies, we are unable to release the raw data. However, we include our evaluation results on this dataset.
\fi
\end{itemize}

\subsection{Metrics}
Following prior work \citep{zhang2021supporting}, we evaluate the quality of the resulting clusters against ground-truth labels using normalized mutual information (NMI) and accuracy. Accuracy is computed by identifying the optimal alignment between predicted clusters and ground-truth labels via the Hungarian algorithm \citep{kuhn1955hungarian}. These metrics are widely adopted in clustering evaluation as they capture both distributional similarity and label-level correspondence.

\subsection{Hyper-parameters and design choices}
Following prior work, we assume the number of clusters $K$ is known \citep{zhang2023clusterllm, viswanathan2024large}. During the embedding-guided subset partition stage, we set the number of groups $M$ to $2K$, enabling more diverse sampling across the dataset. In general, choosing $M > K$ is recommended as it enhances diversity in the sampled records. In the topic extraction stage, we then explicitly prompt the LLM to generate exactly $K$ topics.

The sample size $S$ used for topic extraction is bounded by the LLM's context window. In our experiments this bound is not binding for Bank77, CLINC, Tweet, and Codex, so we use the full dataset ($S=N$); for the larger Adobe Lightroom corpus (6{,}410 records) we set $S=1{,}000$. Assignment always covers every record in $D_N$ regardless of $S$.

To compare effectively with these approaches, we use the same base encoders reported for SCCL, ClusterLLM, Keyphrase Clustering and LLMEdgeRefine: INSTRUCTOR \citep{su2022one} for Bank77 and CLINC, and DistilBERT\footnote{Following ClusterLLM and Keyphrase Clustering, we used a version of DistilBERT finetuned for sentence similarity classification.} \citep{sanh2019distilbert, reimers2019sentence} for Tweet. For our newly constructed domain-specific datasets, we adopt the INSTRUCTOR model.

For the LLM component, we use GPT-4o \citep{hurst2024gpt} as the backbone model for both topic extraction and topic assignment. All GPT-4o calls use the model's default parameters, including its default temperature and token settings. To examine the generalizability of our framework across different LLMs, we additionally evaluate GPT-5 \citep{OpenAI2025GPT5} and Qwen3 (Qwen3-VL-30B-A3B-Instruct) \citep{yang2025qwen3} on the Codex dataset with default settings. To thoroughly evaluate performance, we run each configuration 5 times and report the mean and standard deviation.

\begin{table*}[!t]
  \centering
  \small
  \setlength{\tabcolsep}{4pt}
  \begin{tabular}{@{}lcccccc@{}}
    \toprule
    \multirow{2}{*}{\textbf{Method}} &
    \multicolumn{2}{c}{\textbf{Bank77}} &
    \multicolumn{2}{c}{\textbf{CLINC}} &
    \multicolumn{2}{c}{\textbf{Tweet}} \\
    \cmidrule(lr){2-3} \cmidrule(lr){4-5} \cmidrule(lr){6-7}
     & \textbf{Acc} & \textbf{NMI} & \textbf{Acc} & \textbf{NMI} & \textbf{Acc} & \textbf{NMI} \\
    \midrule

    KMeans                              & 65.2$\pm$0.0 & 83.3$\pm$0.0 & 80.9$\pm$0.0 & 92.5$\pm$0.0 & 59.1$\pm$0.0 & 80.6$\pm$0.0 \\
    Agglomerative                & 64.0$\pm$0.0 & 81.7$\pm$0.0 & 77.7$\pm$0.0 & 91.5$\pm$0.0 & 57.5$\pm$0.0 & 80.6$\pm$0.0 \\
    DSE (w/ KMeans)                    & 49.2$\pm$0.0 & 71.0$\pm$0.0 & 62.1$\pm$0.0 & 85.5$\pm$0.0 & 50.8$\pm$0.0 & 78.4$\pm$0.0 \\
    PCKMeans                            & 59.6$\pm$0.0 & 79.8$\pm$0.0 & 74.5$\pm$0.0 & 90.4$\pm$0.0 & 63.8$\pm$0.0 & 86.7$\pm$0.0 \\
    LLM Correction                      & 64.8    & 80.6    & 75.7    & 91.1    & 67.3    & 87.5 \\
    Keyphrase Clust. (w/ KMeans)        & 67.3$\pm$2.0 & \textbf{83.4$\pm$0.6} & 84.1$\pm$0.5 & 93.1$\pm$0.3 & 61.2$\pm$0.6 & 83.6$\pm$0.9 \\
    Keyphrase Clust. (w/ Agglom.)       & 62.5$\pm$2.0 & 78.2$\pm$0.7 & 81.8$\pm$0.3 & 91.3$\pm$0.1 & 61.8$\pm$0.2 & 84.0$\pm$0.2 \\
    SCCL           & 65.5      & 81.8      & 80.9    & 92.9    & 78.2    & 89.2 \\
    ClusterLLM     & 65.3$\pm$1.8    &  83.3$\pm$0.1      &  79.9$\pm$0.6    &  92.4$\pm$0.4    &  66.3$\pm$1.8      &  88.1$\pm$0.6   \\
    LLMEdgeRefine                      & --      & --      & 86.8    & \textbf{94.9} & --      & --   \\
    \midrule
    \ourmethod (Unsorted)       & 62.7$\pm$1.2 & 80.8$\pm$0.5 & 77.6$\pm$1.7 & 91.2$\pm$0.8 & 71.6$\pm$1.8 & 86.1$\pm$1.0 \\
    \ourmethod (KMeans order)   & 66.3$\pm$3.2 & 82.9$\pm$0.9 & 83.8$\pm$3.9 & 93.9$\pm$1.0 & 84.2$\pm$3.9 & 90.9$\pm$2.5 \\
    \ourmethod (Cosine order)   & \textbf{68.3$\pm$2.9} & 81.4$\pm$0.9 & \textbf{87.3$\pm$1.0} & 93.5$\pm$0.4 & \textbf{87.1$\pm$8.6} & \textbf{91.4$\pm$4.5} \\
    \midrule\midrule
    \ourmethod \textit{(Oracle order)}  & 79.4$\pm$1.3 & 87.2$\pm$0.8 & 90.3$\pm$2.6 & 95.4$\pm$0.6 & 91.3$\pm$3.9 & 94.7$\pm$2.0 \\
    \ourmethod \textit{(Assignment only)}   & 88.8$\pm$0.0 & 92.6$\pm$0.0 & 91.0$\pm$0.0 & 96.6$\pm$0.0 & 92.1$\pm$0.0 & 94.6$\pm$0.0 \\
    \bottomrule
  \end{tabular}
\caption{Clustering performance on three public benchmarks. All LLM-based methods use GPT-4o, except LLMEdgeRefine (GPT-3.5, as we were unable to reproduce the original implementation). Unsorted randomizes input order; Cosine order sorts by cosine similarity (preferred over Euclidean in most cases). Oracle order sorts by ground-truth topics to test the upper bound; Assignment only provides ground-truth topics directly, isolating the assignment stage.}

  \label{tab:results}
\end{table*}

\begin{table*}[t]
  \centering
  \small
  \begin{tabular}{lcccc}
    \toprule
    \multirow{2}{*}{\textbf{Method}} &
    \multicolumn{2}{c}{\textbf{OpenAI Codex}} &
    \multicolumn{2}{c}{\textbf{Adobe Lightroom}} \\
    \cmidrule(lr){2-3} \cmidrule(lr){4-5}
     & \textbf{Acc} & \textbf{NMI} & \textbf{Acc} & \textbf{NMI} \\
    \midrule
    KMeans                              & 41.7$\pm$0.0 & 55.5$\pm$0.0 & 30.7$\pm$0.0 & 18.8$\pm$0.0 \\
    Agglomerative Clust.                & 44.5$\pm$0.0 & 56.1$\pm$0.0 & 16.3$\pm$0.0 & 18.5$\pm$0.0 \\
    Keyphrase Clust. (w/ KMeans)        & 40.3$\pm$2.5 & 52.7$\pm$1.4 & 71.5$\pm$0.3 & 46.8$\pm$0.6 \\
    Keyphrase Clust. (w/ Agglom.)       & 35.8$\pm$1.8 & 48.9$\pm$2.0 & 71.3$\pm$0.4 & 46.0$\pm$0.6 \\
    \midrule
    \ourmethod (Unsorted)            & 61.8$\pm$2.9 & 70.5$\pm$0.9 & 81.9$\pm$2.3 & 56.7$\pm$5.2 \\
    \ourmethod (KMeans order)        & 63.6$\pm$5.0 & 71.5$\pm$3.5 & 82.8$\pm$1.2 & 58.0$\pm$2.6 \\
    \ourmethod (Cosine order)        & \textbf{66.0$\pm$4.6} & \textbf{72.6$\pm$1.8} & \textbf{83.5$\pm$1.9} & \textbf{60.0$\pm$3.2} \\
    \midrule\midrule
    \ourmethod \textit{(Oracle order)}        & 69.9$\pm$3.9 & 75.4$\pm$2.9 & 84.2$\pm$2.1 & 61.6$\pm$2.6 \\
    \ourmethod \textit{(Assignment only)}        & 80.0$\pm$0.0 & 81.6$\pm$0.0 & 92.3$\pm$0.0 & 78.1$\pm$0.0 \\
    \bottomrule
  \end{tabular}
  \caption{Clustering performance on two domain-specific datasets. All LLM based methods use GPT-4o.}
  \label{tab:domain_results}
\end{table*}

\section{Results}
We evaluate \ourmethod on three widely used benchmarks (Bank77, CLINC, Tweet) and two new domain-specific datasets (OpenAI Codex, Adobe Lightroom). We compare \ourmethod against strong baselines and state-of-the-art methods, including DSE \citep{zhou2022learning}, PCKMeans, LLM Correction, and Keyphrase Clustering \citep{viswanathan2024large}, SCCL \citep{zhang2021supporting}, ClusterLLM \citep{zhang2023clusterllm}, and LLMEdgeRefine \citep{feng2024llmedgerefine}.

\subsection{Clustering performance comparison}
Across all five datasets (Tables~\ref{tab:results} and \ref{tab:domain_results}), \ourmethod consistently achieves the highest accuracy, with gains especially pronounced on domain-specific data: on OpenAI Codex, accuracy improves from 44.5 to 66.0 (48\% relative increase). NMI results are similarly strong, with substantial boosts on OpenAI Codex (56.1 to 72.6, 29\% improvement) and Adobe Lightroom (46.8 to 60.0, 28\% improvement).

\subsection{Beyond clustering accuracy: more accurate interpretations}
Standard clustering metrics, accuracy and NMI, assess only whether records are grouped correctly, not whether the resulting clusters are described correctly. We therefore introduce \emph{interpretation accuracy}, measuring how closely a method's cluster label matches the concept it should name: the cosine similarity between the label and the Hungarian-matched ground-truth label \citep{kuhn1955hungarian}, computed with \texttt{all-mpnet-base-v2} embeddings \citep{reimers2019sentence}.

\begin{table}[t]
  \centering
  \small
  \begin{tabular}{@{}lc@{}}
    \toprule
    \textbf{Method} & \textbf{Interpretation Accuracy} \\
    \midrule
    KMeans                     & 0.28 $\pm$ 0.01 \\
    Agglomerative               & 0.27 $\pm$ 0.00 \\
    Keyphrase (w/ KMeans)      & 0.29 $\pm$ 0.02 \\
    Keyphrase (w/ Agglom.)      & 0.28 $\pm$ 0.02 \\
    \midrule
    \ourmethod (Unsorted)       & 0.49 $\pm$ 0.02 \\
    \ourmethod (KMeans order)  & 0.51 $\pm$ 0.06 \\
    \ourmethod (Cosine order)   & \textbf{0.52 $\pm$ 0.03} \\
    \bottomrule
  \end{tabular}
  \caption{Interpretation accuracy on the OpenAI Codex dataset.}
  \label{tab:interp_accuracy}
\end{table}

On OpenAI Codex, \ourmethod attains an interpretation accuracy of 0.52, against 0.27--0.29 for embedding-based methods (Table~\ref{tab:interp_accuracy}). This gap is structural. Because \ourmethod extracts a topic definition and then assigns to it, every cluster is organized around an explicit description; embedding methods cluster first and attach a label only afterward, so their labels describe a group they did not shape. The advantage is intrinsic to the pipeline rather than to any dataset or ordering: the score stays between 0.49 and 0.52 across all input orderings we test.

On top of this structural advantage, the prompt-based interface lets \ourmethod sharpen its definitions with information embeddings cannot exploit (Appendix~\ref{appendix:qualitative}). Given domain knowledge, it separates a competitor mention (``Anthropic is in trouble'') from the generic negative sentiment (``so bad'') that KMeans conflates; given a user preference, it consolidates off-topic presentation remarks (a salad bowl, the presenter's beard) into a single low-priority cluster instead of fragmenting the fixed cluster budget as Keyphrase does. We quantify the accuracy gains from these interventions in \S\ref{sec:ablation}.

\subsection{Ablation study}
\label{sec:ablation}

\paragraph{Topic extraction limits performance.}
To disentangle the contributions of topic extraction and topic assignment, we skip the topic extraction stage and directly provide ground-truth topics. As shown in Tables~\ref{tab:results} and \ref{tab:domain_results}, performance becomes nearly saturated (accuracy and NMI exceed 85 points on generic benchmarks and are around 80--90 on domain-specific datasets), with extremely small variance. This indicates that the assignment stage is already highly reliable, and the primary opportunity for improvement lies in topic extraction.

\begin{figure*}[t]
  \centering
  \includegraphics[width=\textwidth]{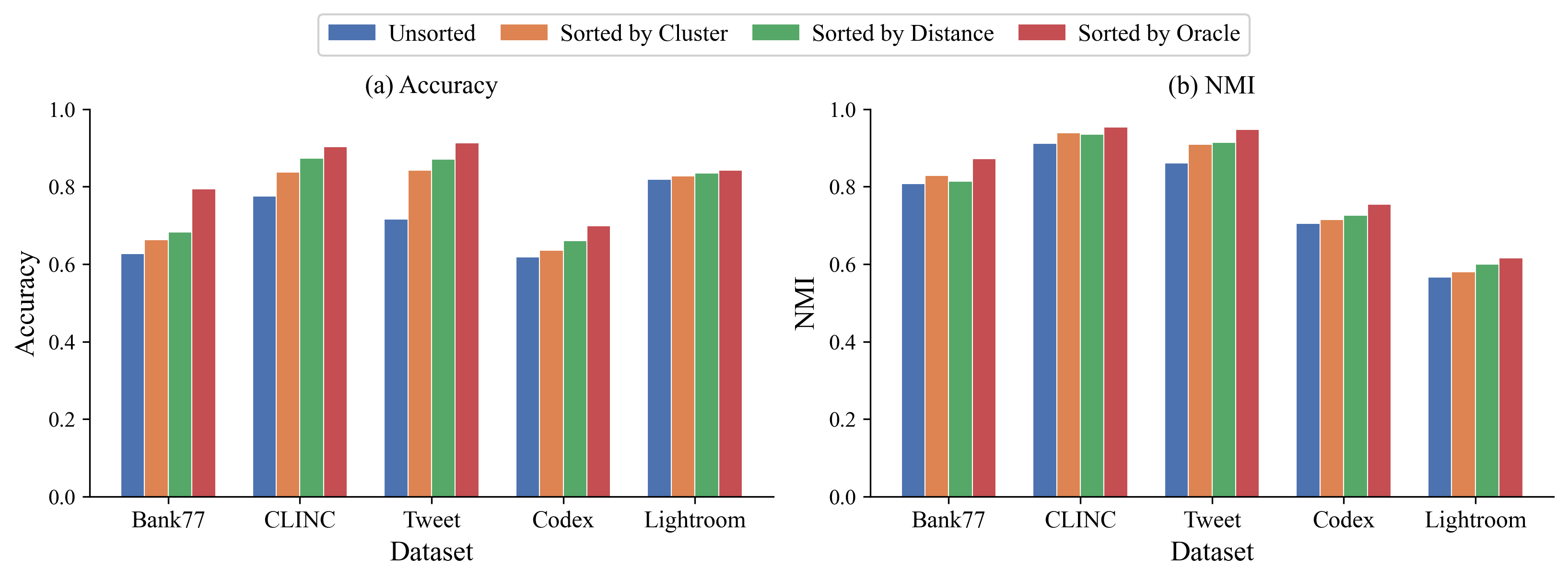}
  \caption{Clustering performance under different ordering strategies across five datasets: (a) accuracy and (b) NMI.}
  \label{fig:ordering}
\end{figure*}

\paragraph{Ordering matters to topic extraction.}
During our experiments, we observed that exemplar ordering plays a critical role in the quality of topic extraction. As shown in Table~\ref{tab:domain_results}, even the unsorted baseline already outperforms traditional clustering methods. However, performance improves consistently when records are pre-organized---whether by cluster assignments (e.g., KMeans) or by embedding similarity (e.g., cosine distance)---as illustrated in Figure~\ref{fig:ordering}. On the Tweet dataset, for instance, accuracy rises by 22\% (from 71.6 to 87.1) solely due to improved ordering.

One explanation lies in how transformer-based LLMs handle long contexts. While self-attention allows tokens to interact globally, empirical studies show that attention effectiveness decays with distance, making it harder to integrate scattered signals \citep{press2021train, liu2023lost}. Pre-organizing records into semantically coherent groups reduces this burden and allows the LLM to better exploit its contextual reasoning. We further validate this with an oracle ordering (sorted by ground-truth topics): while infeasible in practice, the consistent gap between heuristic and oracle orderings confirms that improved ordering heuristics could unlock further gains.

\paragraph{The pipeline alone outperforms baselines; domain prompting adds further gains.}
We disentangle \ourmethod's gains with an ablation on extra domain prompting (Table~\ref{tab:prompting_ablation}). The pipeline alone is already decisive. With no extra domain prompting, using only the minimal context available to baselines, \ourmethod achieves the best accuracy on all five datasets, by a wide margin on the domain-specific corpora (Codex 53.8 vs.\ 44.5; Lightroom 78.2 vs.\ 71.5). This advantage is structural and needs no extra domain input. Additional domain prompting then adds a further boost. Because the LLM is the clustering core, \ourmethod can naturally absorb background knowledge and user preferences through prompting. This boost is large where the LLM lacks domain knowledge (Codex 53.8$\rightarrow$66.0; Lightroom 78.2$\rightarrow$83.5) and negligible on the general benchmarks.\footnote{Even a general benchmark can benefit from a preference that encodes ground-truth granularity: on CLINC, one instruction to treat ambiguous responses (``yes''/``no''/``unsure'') as separate intents rather than one group raises accuracy from 87.1 to 87.3.} The two gains are additive: on Codex, the pipeline contributes $+9.3$ points and prompting a further $+12.2$. It is also important to note that the extra prompting stays high-level. It supplies background knowledge and coarse organizing preferences (e.g., that presentation-related comments should form one cluster). It never enumerates the target clusters or assigns any record. The model still extracts the topics from the data and labels each record itself. For full transparency, we provide the exact prompts we use for Codex in Appendix~\ref{appendix:domainknowledge_prompt}.

\begin{table*}[t]
  \centering
  \small
  \setlength{\tabcolsep}{4pt}
  \begin{tabular}{@{}lccccc@{}}
    \toprule
    \textbf{Method} & \textbf{Bank77} & \textbf{CLINC} & \textbf{Tweet} & \textbf{OpenAI Codex} & \textbf{Adobe Lightroom} \\
    \midrule
    Best baseline & 67.3$\pm$2.0 & 86.8 & 78.2 & 44.5$\pm$0.0 & 71.5$\pm$0.3 \\
    \ourmethod (w/o extra domain prompting) & \textbf{68.3$\pm$2.9} & 87.1$\pm$0.8 & \textbf{87.1$\pm$8.6} & 53.8$\pm$2.0 & 78.2$\pm$2.1 \\
    \ourmethod (w/ extra domain prompting) & -- & \textbf{87.3$\pm$1.0} & -- & \textbf{66.0$\pm$4.6} & \textbf{83.5$\pm$1.9} \\
    \bottomrule
  \end{tabular}
  \caption{Accuracy with and without extra domain prompting. Using only the minimal context available to baselines (w/o extra), \ourmethod already achieves the best accuracy on all five datasets; extra domain prompting adds a further boost on the domain-specific corpora (Codex, Lightroom). ``--'': no extra domain prompting was applied, as none was found to improve these general benchmarks.}
  \label{tab:prompting_ablation}
\end{table*}

% Table (LLM backbones) moved to Appendix~\ref{appendix:additional_results} for the page-limit.

\paragraph{Best performance across LLM families}

To evaluate generalizability, we additionally test GPT-5 \citep{OpenAI2025GPT5} and Qwen3 \citep{yang2025qwen3} on the Codex dataset. As shown in Table~\ref{tab:codex_llm_backbones} (Appendix~\ref{appendix:additional_results}), our method remains the best-performing approach across all model families, with ordering consistently providing a substantial boost. The ``Assignment Only'' setting varies little across LLMs, while performance differences are more pronounced in topic extraction, reinforcing that it is the primary challenge.

\subsection{Cost performance}
While clustering accuracy is critical, practical deployment also requires methods to be cost-efficient. Unlike representation-learning approaches such as SCCL \citep{zhang2021supporting} or DSE \citep{zhou2022learning}, \ourmethod is training-free: it requires no fine-tuning or GPU-intensive optimization, relying only on one-off clustering and prompting.

Compared to other training-free, LLM-based methods, \ourmethod incurs a cost profile similar to the Keyphrase method \citep{viswanathan2024large}, and its two stages scale complementarily. Topic extraction issues a single LLM call over a fixed-size sample of $S$ records, so its cost is independent of the corpus size $N$; topic assignment issues one call per record, so the dominant cost is linear in $N$ and, because records are labeled independently, trivially parallelizable. \ourmethod thus scales to large corpora at the asymptotic cost of a single per-record LLM pass, and when the representative sample exceeds the context window, topic extraction extends naturally to a batched variant (see Limitations). Crucially, at comparable cost, \ourmethod delivers substantially higher accuracy, demonstrating superior cost-effectiveness.

\begin{figure}[t]
  \centering
  \includegraphics[width=\linewidth]{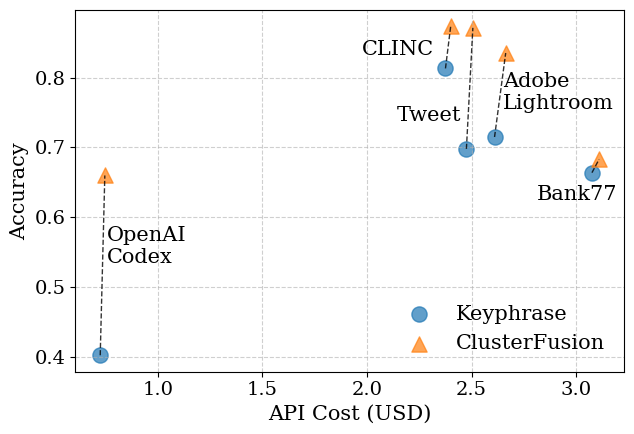}
  \caption{Cost-accuracy comparison between \ourmethod and Keyphrase. API price is based on OpenAI's GPT-4o pricing (April 2025).}
  \label{fig:cost_accuracy}
\end{figure}

As shown in Figure~\ref{fig:cost_accuracy}, \ourmethod consistently achieves higher accuracy at nearly identical cost across all five datasets, with particularly pronounced gains on OpenAI Codex (a 63.8\% relative accuracy improvement over Keyphrase with less than \$0.02 additional cost); the monetary cost we report is computed from token-based API pricing and therefore directly reflects token usage.

\section{Conclusion}
We recast text clustering with the LLM as its core: rather than partitioning an embedding space and labeling the result after the fact, \ourmethod extracts explicit topic definitions and assigns each record by conditioning on them. This renders the clusters interpretable by construction, and we further showed that the extracted definitions are accurate descriptions of the underlying concepts, a quality embedding-based partitions cannot provide. Analyzing the paradigm, we found that its difficulty lies not in assignment, which is near-saturated, but in topic extraction, whose quality is governed by \emph{context organization}: reordering the same inputs alone shifts accuracy by up to 22\%. On this basis, \ourmethod adapts to specialized domains through prompting alone, without any training, and attains the best accuracy across five datasets. Together, these results establish LLM-core clustering as an accurate and interpretable alternative to embedding-based pipelines, and identify context organization as the central lever for improving it further.

\section*{Limitations}
While \ourmethod achieves strong performance, several limitations remain. First, for extremely large datasets, our embedding-guided partition naturally extends to batch processing: embeddings can organize samples both within each batch (via ordering) and across batches (via balanced sampling), enabling hierarchical or parallel topic extraction while preserving the framework's core advantages. Second, our main experiments assume the number of clusters $K$ is known \citep{zhang2023clusterllm, viswanathan2024large, zhang2021supporting, feng2024llmedgerefine}. This is a deliberate design choice to enable fair comparison with prior work, not a constraint of our method. Because topics are extracted by the LLM, the point value of $K$ can be replaced with a user-specified range $[K_{\min}, K_{\max}]$ from which the model selects. This requires only a one-line change to the topic extraction prompt, and captures a user's intended granularity better than a brittle point estimate. This range variant performs strongly in our internal experiments; we report fixed-$K$ throughout only because these baselines all require a known $K$, and adapting them for an apples-to-apples range-based comparison is beyond our scope. A systematic study of range-based and automatic $K$ selection is a natural next step.

\section*{Ethics Statement}
\paragraph{Data privacy.}
We use public benchmarks (BANK77, CLINC, Tweet) and two newly constructed ones based on public Adobe Lightroom and Codex reviews. All sources are public and free of personally identifiable information (PII), and comply with their respective terms of use. No sensitive or proprietary data is involved.

\paragraph{Bias in LLMs.}
Our method uses GPT-4o \citep{hurst2024gpt} for topic extraction and assignment, which may reflect biases from its training data. While our balanced sampling strategy helps mitigate this, we acknowledge potential societal or linguistic bias. Future work could explore fairness-aware prompting or bias auditing.

\paragraph{Reproducibility.}
We document each step of our pipeline in detail. Grouping uses open-source libraries (such as KMeans), and LLM-based steps rely on modular, prompt-based methods. We also release the dataset and raw experimental results to support reproducibility.

\paragraph{Environmental impact.}
To reduce compute cost, we limit LLM usage through lightweight grouping and representative sampling. Nonetheless, we recognize the environmental footprint of LLM inference and encourage future work on more efficient alternatives.

% Acknowledgments must appear immediately before the references and are omitted
% from the anonymous review version (hidden when \ifsubmission is true).
\ifsubmission
\else
\section*{Acknowledgments}
This work was conducted under the support of the Adobe Experience Intelligence team. We would like to thank Lei Liu, Austin Xu, Pradeep Arockiasamy, William Yan and Roger Brooks for their valuable feedback and support throughout the development of this work.
\fi

% The \bibliographystyle is set to acl_natbib by the acl package, so we only
% need to point at the bibliography database here.
\bibliography{custom}

@article{zhang2021supporting,
  title={Supporting clustering with contrastive learning},
  author={Zhang, Dejiao and Nan, Feng and Wei, Xiaokai and Li, Shangwen and Zhu, Henghui and McKeown, Kathleen and Nallapati, Ramesh and Arnold, Andrew and Xiang, Bing},
  journal={arXiv preprint arXiv:2103.12953},
  year={2021}
}

@inproceedings{jo-etal-2025-zerodl,
    title = "{Z}ero{DL}: Zero-shot Distribution Learning for Text Clustering via Large Language Models",
    author = "Jo, Hwiyeol and Lee, Hyunwoo and Yoo, Kang Min and Park, Taiwoo",
    booktitle = "Findings of the Association for Computational Linguistics: ACL 2025",
    year = "2025",
    publisher = "Association for Computational Linguistics",
    url = "https://aclanthology.org/2025.findings-acl.1005/",
    pages = "19597--19607"
}

@inproceedings{huang2025text,
  title={Text clustering as classification with llms},
  author={Huang, Chen and He, Guoxiu},
  booktitle={Proceedings of the 2025 Annual International ACM SIGIR Conference on Research and Development in Information Retrieval in the Asia Pacific Region},
  pages={374--384},
  year={2025}
}

@article{day1984efficient,
  title={Efficient algorithms for agglomerative hierarchical clustering methods},
  author={Day, William HE and Edelsbrunner, Herbert},
  journal={Journal of classification},
  volume={1},
  number={1},
  pages={7--24},
  year={1984},
  publisher={Springer}
}

@article{kuhn1955hungarian,
  title={The Hungarian method for the assignment problem},
  author={Kuhn, Harold W},
  journal={Naval research logistics quarterly},
  volume={2},
  number={1-2},
  pages={83--97},
  year={1955},
  publisher={Wiley Online Library}
}

@article{muennighoff2022mteb,
  title={Mteb: Massive text embedding benchmark},
  author={Muennighoff, Niklas and Tazi, Nouamane and Magne, Lo{\"\i}c and Reimers, Nils},
  journal={arXiv preprint arXiv:2210.07316},
  year={2022}
}

@article{viswanathan2024large,
  title={Large language models enable few-shot clustering},
  author={Viswanathan, Vijay and Gashteovski, Kiril and Gashteovski, Kiril and Lawrence, Carolin and Wu, Tongshuang and Neubig, Graham},
  journal={Transactions of the Association for Computational Linguistics},
  volume={12},
  pages={321--333},
  year={2024},
  publisher={MIT Press One Broadway, 12th Floor, Cambridge, Massachusetts 02142, USA~…}
}

@article{wang2023goal,
  title={Goal-driven explainable clustering via language descriptions},
  author={Wang, Zihan and Shang, Jingbo and Zhong, Ruiqi},
  journal={arXiv preprint arXiv:2305.13749},
  year={2023}
}

@article{de2023idas,
  title={IDAS: Intent discovery with abstractive summarization},
  author={De Raedt, Maarten and Godin, Fr{\'e}deric and Demeester, Thomas and Develder, Chris},
  journal={arXiv preprint arXiv:2305.19783},
  year={2023}
}

@article{bae2020interactive,
author = {Bae, Juhee and Helldin, Tove and Riveiro, Maria and Nowaczyk, S{\l}awomir and Bouguelia, Mohamed-Rafik and Falkman, G{\"o}ran},
title = {Interactive Clustering: A Comprehensive Review},
year = {2020},
issue_date = {January 2021},
publisher = {Association for Computing Machinery},
address = {New York, NY, USA},
volume = {53},
number = {1},
issn = {0360-0300},
url = {https://doi.org/10.1145/3340960},
doi = {10.1145/3340960},
journal = {ACM Comput. Surv.},
month = feb,
articleno = {1},
numpages = {39}
}

@article{larson2019evaluation,
  title={An evaluation dataset for intent classification and out-of-scope prediction},
  author={Larson, Stefan and Mahendran, Anish and Peper, Joseph J and Clarke, Christopher and Lee, Andrew and Hill, Parker and Kummerfeld, Jonathan K and Leach, Kevin and Laurenzano, Michael A and Tang, Lingjia and others},
  journal={arXiv preprint arXiv:1909.02027},
  year={2019}
}

@article{casanueva2020efficient,
  title={Efficient intent detection with dual sentence encoders},
  author={Casanueva, I{\~n}igo and Tem{\v{c}}inas, Tadas and Gerz, Daniela and Henderson, Matthew and Vuli{\'c}, Ivan},
  journal={arXiv preprint arXiv:2003.04807},
  year={2020}
}

@misc{OpenAI2025GPT5,
  author = {{OpenAI}},
  title = {{GPT-5 Large language model}},
  howpublished = {\url{https://openai.com/gpt-5/}},
  year = {2025},
  note = {{Accessed: 19 November 2025}}
}

@article{yang2025qwen3,
  title={Qwen3 technical report},
  author={Yang, An and Li, Anfeng and Yang, Baosong and Zhang, Beichen and Hui, Binyuan and Zheng, Bo and Yu, Bowen and Gao, Chang and Huang, Chengen and Lv, Chenxu and others},
  journal={arXiv preprint arXiv:2505.09388},
  year={2025}
}

@article{hurst2024gpt,
  title={Gpt-4o system card},
  author={Hurst, Aaron and Lerer, Adam and Goucher, Adam P and Perelman, Adam and Ramesh, Aditya and Clark, Aidan and Ostrow, AJ and Welihinda, Akila and Hayes, Alan and Radford, Alec and others},
  journal={arXiv preprint arXiv:2410.21276},
  year={2024}
}

@inproceedings{press2021train,
  title={Train Short, Test Long: Attention with Linear Biases Enables Input Length Extrapolation},
  author={Press, Ofir and Smith, Noah A. and Levy, Mike},
  booktitle={Proceedings of the 59th Annual Meeting of the Association for Computational Linguistics (ACL)},
  year={2021},
  pages={280--291}
}

@article{liu2023lost,
  title={Lost in the Middle: How Language Models Use Long Contexts},
  author={Liu, Nelson F. and Lin, Kevin and Hewitt, John and Paranjape, Ashwin and Bevilacqua, Michele and Mitra, Arjun and Liang, Percy},
  journal={Transactions of the Association for Computational Linguistics},
  volume={11},
  pages={157--173},
  year={2023},
  publisher={MIT Press}
}

@inproceedings{yin2016model,
  title={A model-based approach for text clustering with outlier detection},
  author={Yin, Jianhua and Wang, Jianyong},
  booktitle={2016 IEEE 32nd International Conference on Data Engineering (ICDE)},
  pages={625--636},
  year={2016},
  organization={IEEE}
}

@inproceedings{feng2024llmedgerefine,
  title={LLMEdgeRefine: Enhancing text clustering with LLM-based boundary point refinement},
  author={Feng, Zijin and Lin, Luyang and Wang, Lingzhi and Cheng, Hong and Wong, Kam-Fai},
  booktitle={Proceedings of the 2024 Conference on Empirical Methods in Natural Language Processing},
  pages={18455--18462},
  year={2024}
}

@article{zhou2022learning,
  title={Learning dialogue representations from consecutive utterances},
  author={Zhou, Zhihan and Zhang, Dejiao and Xiao, Wei and Dingwall, Nicholas and Ma, Xiaofei and Arnold, Andrew O and Xiang, Bing},
  journal={arXiv preprint arXiv:2205.13568},
  year={2022}
}

@article{sanh2019distilbert,
  title={DistilBERT, a distilled version of BERT: smaller, faster, cheaper and lighter},
  author={Sanh, Victor and Debut, Lysandre and Chaumond, Julien and Wolf, Thomas},
  journal={arXiv preprint arXiv:1910.01108},
  year={2019}
}

@article{reimers2019sentence,
  title={Sentence-bert: Sentence embeddings using siamese bert-networks},
  author={Reimers, Nils and Gurevych, Iryna},
  journal={arXiv preprint arXiv:1908.10084},
  year={2019}
}

@article{su2022one,
  title={One embedder, any task: Instruction-finetuned text embeddings},
  author={Su, Hongjin and Shi, Weijia and Kasai, Jungo and Wang, Yizhong and Hu, Yushi and Ostendorf, Mari and Yih, Wen-tau and Smith, Noah A and Zettlemoyer, Luke and Yu, Tao},
  journal={arXiv preprint arXiv:2212.09741},
  year={2022}
}

@article{zhang2023clusterllm,
  title={Clusterllm: Large language models as a guide for text clustering},
  author={Zhang, Yuwei and Wang, Zihan and Shang, Jingbo},
  journal={arXiv preprint arXiv:2305.14871},
  year={2023}
}

@inproceedings{basu2004active,
  title={Active semi-supervision for pairwise constrained clustering},
  author={Basu, Sugato and Banerjee, Arindam and Mooney, Raymond J},
  booktitle={Proceedings of the 2004 SIAM international conference on data mining},
  pages={333--344},
  year={2004},
  organization={SIAM}
}

@inproceedings{basu2002semi,
  title={Semi-supervised clustering by seeding},
  author={Basu, Sugato and Banerjee, Arindam and Mooney, Raymond J},
  booktitle={Proceedings of the nineteenth international conference on machine learning},
  pages={27--34},
  year={2002}
}

@article{wang2022text,
  title={Text embeddings by weakly-supervised contrastive pre-training},
  author={Wang, Liang and Yang, Nan and Huang, Xiaolong and Jiao, Binxing and Yang, Linjun and Jiang, Daxin and Majumder, Rangan and Wei, Furu},
  journal={arXiv preprint arXiv:2212.03533},
  year={2022}
}

@inproceedings{MacQueen1967,
  author = {MacQueen, J. B.},
  booktitle = {Proc. of the fifth Berkeley Symposium on Mathematical Statistics and Probability},
  editor = {Cam, L. M. Le and Neyman, J.},
  pages = {281-297},
  publisher = {University of California Press},
  title = {Some Methods for Classification and Analysis of MultiVariate Observations},
  volume = 1,
  year = 1967
}

@inproceedings{Caruana-2013,
  author = {Caruana, Rich},
  title = {Clustering: probably approximately useless?},
  year = {2013},
  isbn = {9781450322638},
  publisher = {Association for Computing Machinery},
  address = {New York, NY, USA},
  url = {https://doi.org/10.1145/2505515.2514692},
  doi = {10.1145/2505515.2514692},
  booktitle = {Proceedings of the 22nd ACM International Conference on Information \& Knowledge Management},
  pages = {1259--1260},
  numpages = {2},
  location = {San Francisco, California, USA},
  series = {CIKM '13}
}

@article{Johnson2023WhereDM,
  title={Where Does My Model Underperform? A Human Evaluation of Slice Discovery Algorithms},
  author={Nari Johnson and {\'A}ngel Alexander Cabrera and Gregory Plumb and Ameet Talwalkar},
  journal={ArXiv},
  year={2023},
  volume={abs/2306.08167},
  url={https://api.semanticscholar.org/CorpusID:259165223}
}

@article{Charikar1997IncrementalCA,
  title={Incremental clustering and dynamic information retrieval},
  author={Moses Charikar and Chandra Chekuri and Tom{\'a}s Feder and Rajeev Motwani},
  journal={SIAM J. Comput.},
  year={1997},
  volume={33},
  pages={1417-1440},
  url={https://api.semanticscholar.org/CorpusID:6524917}
}

@inproceedings{jacovi-goldberg-2020-towards,
    title = "Towards Faithfully Interpretable {NLP} Systems: How Should We Define and Evaluate Faithfulness?",
    author = "Jacovi, Alon  and
      Goldberg, Yoav",
    editor = "Jurafsky, Dan  and
      Chai, Joyce  and
      Schluter, Natalie  and
      Tetreault, Joel",
    booktitle = "Proceedings of the 58th Annual Meeting of the Association for Computational Linguistics",
    month = jul,
    year = "2020",
    address = "Online",
    publisher = "Association for Computational Linguistics",
    url = "https://aclanthology.org/2020.acl-main.386/",
    doi = "10.18653/v1/2020.acl-main.386",
    pages = "4198--4205"
}

\appendix

\section{Additional experimental results}
\label{appendix:additional_results}

Table~\ref{tab:codex_llm_backbones} reports \ourmethod's performance on the Codex dataset across three LLM backbones (GPT-4o, GPT-5, and Qwen3), supporting the generalizability analysis in \S\ref{sec:ablation}.

\begin{table*}[t]
  \centering
  \small
  \setlength{\tabcolsep}{4pt}
  \begin{tabular}{@{}lcccccc@{}}
    \toprule
    \multirow{2}{*}{\textbf{Method}} &
    \multicolumn{2}{c}{\textbf{GPT-4o}} &
    \multicolumn{2}{c}{\textbf{GPT-5}} &
    \multicolumn{2}{c}{\textbf{Qwen3}} \\
    \cmidrule(lr){2-3} \cmidrule(lr){4-5} \cmidrule(lr){6-7}
     & \textbf{Acc} & \textbf{NMI} & \textbf{Acc} & \textbf{NMI} & \textbf{Acc} & \textbf{NMI} \\
    \midrule
    Keyphrase Clust.\ (w/ KMeans)   & 40.3$\pm$2.5 & 52.7$\pm$1.4 & 40.9$\pm$1.6 & 53.0$\pm$2.3 & 31.2$\pm$3.3 & 45.2$\pm$1.3 \\
    Keyphrase Clust.\ (w/ Agglom.)  & 35.8$\pm$1.8 & 48.9$\pm$2.0 & 36.1$\pm$2.2 & 49.3$\pm$2.7 & 30.1$\pm$1.5 & 40.2$\pm$1.3 \\
    \midrule
    \ourmethod (Unsorted)           & 61.8$\pm$2.9 & 70.5$\pm$0.9 & 64.9$\pm$1.4 & 70.9$\pm$1.3 & 51.6$\pm$2.9 & 63.6$\pm$1.4 \\
    \ourmethod (KMeans order)       & 63.6$\pm$5.0 & 71.5$\pm$3.5 & 65.3$\pm$6.9 & 75.4$\pm$4.0 & 55.1$\pm$1.4 & 65.3$\pm$1.7 \\
    \ourmethod (Cosine order)       & \textbf{66.0$\pm$4.6} & \textbf{72.6$\pm$1.8} & \textbf{69.0$\pm$3.5} & \textbf{76.5$\pm$1.4} & \textbf{56.0$\pm$2.5} & \textbf{66.6$\pm$1.7} \\
    \midrule\midrule
    \ourmethod \textit{(Oracle order)}        & 69.9$\pm$3.9 & 75.4$\pm$2.9 & 80.0$\pm$4.5 & 82.0$\pm$2.2 & 61.4$\pm$5.6 & 68.1$\pm$4.1 \\
    \ourmethod \textit{(Assignment only)}     & 80.0$\pm$0.0 & 81.6$\pm$0.0 & 80.5$\pm$0.0 & 81.4$\pm$0.0 & 78.1$\pm$0.0 & 79.4$\pm$0.0 \\
    \bottomrule
  \end{tabular}
  \caption{Clustering performance on the Codex dataset across different LLM backbones.}
  \label{tab:codex_llm_backbones}
\end{table*}

\section{LLM prompts}
\label{appendix:prompts}

This appendix provides the complete prompt templates used in our \ourmethod framework for both topic extraction and topic assignment stages.

\subsection{Topic extraction prompt}
\label{appendix:summarization_prompt}

The following prompt template is used to extract candidate topics from the sampled and sorted records. The prompt follows a two-message structure:

\begin{quote}
  \small
  \textbf{System message:} \textit{You are an intelligent assistant skilled in summarizing and extracting insights.}

  \textbf{User message:} \textit{You are now tasked with reviewing records related to \{feature\_context\}. Each record is separated by ``;'' symbols.}

  \textit{Records: \{concatenated\_records\}}

  \textit{Your goal is to extract the key topics from these records. For each identified topic, please provide a brief explanation along with examples. The total number of topics you extract should be exactly \{num\_topics\}, not more than \{num\_topics\} and not fewer than \{num\_topics\}.}

  \textit{The result should be returned in JSON format, where each key represents an index, and the corresponding value is a dictionary with:}
  \begin{itemize}
  \item \textit{A topic name as the key, and}
  \item \textit{A description and some examples as the value.}
  \end{itemize}

  \textit{\{extra\_guidance if extra\_guidance else ``''\}}
\end{quote}

where \texttt{feature\_context} specifies the domain or context of the dataset, \texttt{concatenated\_records} contains the sampled and sorted records from $D_S'$ (as generated in Equation 1-5) concatenated with ``;'' separators, \texttt{num\_topics} is the target number of clusters $K$, and the optional \texttt{extra\_guidance} placeholder can be used to provide additional user preference when needed.

\subsection{Domain-specific adaptations}
\label{appendix:domainknowledge_prompt}
For domain-specific datasets, we provide task-specific context and preferences to guide topic formation. Below we show the adaptations for the OpenAI Codex dataset as an illustrative example:

\paragraph{Feature context (domain knowledge):}
\begin{quote}
\small
\textit{OpenAI Codex feature announcement video posted on YouTube. Comments are from viewers responding to the demo by Fouad Matin and Romain Huet, showing Codex CLI working with OpenAI models including o3, o4-mini, and GPT-4.1. In this domain,  Manus, Aider, Deepseek, Claude (by Anthropic), Cursor, Warp are competitors; Gemini and Deepseek are non-OpenAI models. There are also certain YouTube channels making tech tutorial/review videos, such as Fireship and Lucas Montano. Notably, the recorded demo had some staging details: a salad bowl placed on the table, the interface displayed in light mode, and a smiley-face sticker covering the Apple logo on the MacBook.}
\end{quote}

\paragraph{Extra guidance (user preferences):}
\begin{quote}
\small
\textit{Special instructions for topic formation:}
\begin{itemize}
\item \textit{If the comment is not related to codex feature but related to presentation or presenters (such as salad bowl, MacBook sticker), consolidate them into one cluster}
\end{itemize}
\end{quote}

These adaptations directly enable the qualitative improvements shown in Table~\ref{tab:qualitative_examples}, where domain terminology is correctly interpreted and non-substantive feedback is consolidated. The complete prompts for all datasets will be released on GitHub to support reproducibility.

\subsection{Topic assignment prompt}
\label{appendix:assignment_prompt}

For classifying each record into one of the extracted topics, we use the following two-message prompt structure:

\begin{quote}
\small
\textbf{System message:} \textit{You are a helpful assistant, that can help me label each record into topics.}

\textbf{User message:} \textit{Following records is about \{feature\_context\}. Please classify the record into one of the following topics, which are represented as a dictionary. Its keys are the names of the topics and values are the descriptions of the topic: \{topic\_def\_dict\}}

\textit{Comment: \{comment\}}

\textit{Return the result in JSON format with the following format: key `topic', with value as the picked topic;}
\end{quote}

where \texttt{topic\_def\_dict} contains the topics and descriptions generated in the topic extraction stage, and \texttt{comment} is the individual record being classified.

\section{Qualitative examples}
\label{appendix:qualitative}

\begin{table*}[t]
\centering
\small
\setlength{\tabcolsep}{2pt}
\begin{tabular}{@{}p{2.7cm}p{2.4cm}p{2.4cm}p{2.7cm}p{2.6cm}@{}}
    \toprule
    \textbf{Comment} & \textbf{Ground truth} & \textbf{KMeans} & \textbf{Keyphrase} & \textbf{\ourmethod} \\
    \midrule
    \multicolumn{5}{l}{\textit{Domain Knowledge: Understand Emerging Tech Terminology}} \\
    \midrule
    ``Anthropic is in trouble'' & Competition With Competitors & Pessimistic Expectations  & Impact of AI on Employment and Software Development & Competitors Emergence \\
    ``so bad'' & General Criticism & Pessimistic Expectations & General Criticism & General Criticism \\
    ``someone just needs to come up with a way to run 670b deepseek models, so we can get rid of the cost'' & Profit Model Discussion & Cost and lack of local implementation & Competitor solutions outperforming current offerings & Better solutions provided by competitors \\
    ``Please automate my mortgage and bills'' & Future Use Case & Cost and lack of local implementation & Technology Ethics, Privacy, and Practical Applications & End-user Feature Requests \\
    \midrule
    \multicolumn{5}{l}{\textit{User Preference: Consolidate Non-Feature Related Comments}} \\
    \midrule
    ``Are those strawberries and broccoli in the bowl?'' & Feedback On The Presenter And Presentation Style & Table items and food discussions & Strawberry-related inquiries & Presenter Criticism and Observations \\
    ``that random salad decoration lol'' & Feedback On The Presenter And Presentation Style & Table items and food discussions & Video Production and Presentation Critiques & Presenter Criticism and Observations \\
    ``Interesting beard choice'' & Feedback On The Presenter And Presentation Style & Excitement and Enthusiasm & Presenter Appearance and Personality Observations & Presenter Criticism and Observations \\
    ``omg he hit the corner on the Mac on the table, hahahaha'' & Feedback On The Presenter And Presentation Style & Cursor Tool and Presentation Observations & Humorous observations about MacBook setup and sticker usage & Presenter Criticism and Observations \\
    \bottomrule
\end{tabular}
\caption{Qualitative examples from the OpenAI Codex dataset.}
\label{tab:qualitative_examples}
\end{table*}

\end{document}